\documentclass[journal]{IEEEtran}
\IEEEoverridecommandlockouts

\usepackage{cite}
\usepackage{amsmath,amssymb,amsfonts}
\usepackage{algorithmic}
\usepackage{graphicx}
\usepackage{textcomp}
\usepackage{xcolor}
\usepackage{float}
\usepackage{overpic}
\usepackage{multirow}
\usepackage{subfigure}
\usepackage[colorlinks,
linkcolor=red,
anchorcolor=red,
citecolor=green]{hyperref}
\def\BibTeX{{\rm B\kern-.05em{\sc i\kern-.025em b}\kern-.08em
    T\kern-.1667em\lower.7ex\hbox{E}\kern-.125emX}}

\begin{document}

	\title{Deep Pneumonia: Attention-Based Contrastive Learning for Class-Imbalanced Pneumonia Lesion Recognition in Chest X-rays 	\thanks{This work was supported by Guangdong Key R\&D Project (\#2018B030338001) and Natural Science Foundations of China (\#61806041, \#62076055). (Corresponding author: Xian-Shi~Zhang, email: zhangxianshi@uestc.edu.cn)}%
		\thanks{Xinxu~Wei, Haohan Bai, Xianshi~Zhang 
			and Yongjie~Li are with the MOE Key Lab for Neuroinformation, University of Electronic Science and Technology of China (UESTC), Chengdu 610054, China.
		Xiangke Niu is with Department of Radiology, Affiliated Hospital of Chengdu University.}}

	\author{Xinxu~Wei, Xiangke Niu, Xianshi Zhang\IEEEauthorrefmark{1}, 
	and Yongjie~Li,~\IEEEmembership{Senior Member,~IEEE}}




	\maketitle


\maketitle

\begin{abstract}
Computer-aided X-ray pneumonia lesion recognition is important for accurate diagnosis of pneumonia. With the emergence of deep learning, the identification accuracy of pneumonia has been greatly improved, but there are still some challenges due to the fuzzy appearance of chest X-rays. In this paper, we propose a deep learning framework named Attention-Based Contrastive Learning for Class-Imbalanced X-Ray Pneumonia Lesion Recognition (denoted as Deep Pneumonia). We adopt self-supervised contrastive learning strategy to pre-train the model without using extra pneumonia data for fully mining the limited available dataset. In order to leverage the location information of the lesion area that the doctor has painstakingly marked, we propose mask-guided hard attention strategy and feature learning with contrastive regulation strategy which are applied on the attention map and the extracted features respectively to guide the model to focus more attention on the lesion area where contains more discriminative features for improving the recognition performance. In addition, we adopt Class-Balanced Loss instead of traditional Cross-Entropy as the loss function of classification to tackle the problem of serious class imbalance between different classes of pneumonia in the dataset. The experimental results show that our proposed framework can be used as a reliable computer-aided pneumonia diagnosis system to assist doctors to better diagnose pneumonia cases accurately. 
\end{abstract}


\section{Introduction}
Pneumonia is usually caused by the viral or bacterial infection. Pneumonia is often diagnosed based on symptoms as well as physical examination. Early diagnosis of pneumonia is very important for effective treatment of pneumonia. Chest X-rays are commonly used as a clinical method to diagnose pneumonia. Chest X-ray radiograph usually requires the examine by highly-trained radiologists, and usually there exists disagreement among these experienced experts \cite{rajpurkar2017chexnet}. 
X-ray radiographs of chest pneumonia are usually very fuzzy, and there are other kinds of lesions in both lungs, such as lung cancer and excess fluid, which show similar opacities with pneumonia lesion in X-ray images, making it difficult to diagnose pneumonia and locate the area of the pneumonia lesion \cite{gabruseva2020deep} \cite{jaiswal2019identifying}.
Other diagnostic methods can help confirm the diagnosis, such as blood tests, and sputum microbial cultures. But these diagnostic methods are time-consuming and costly. In contrast, diagnostic methods aided by computer systems and artificial intelligence algorithms have the advantages of being fast, convenient and accurate. 
In recent years, deep learning has achieved breakthrough success in many fields, such as medical image recognition\cite{trivedy2020design}\cite{hao2021fusing}, object detection\cite{girshick2015fast}\cite{xie2019detecting} and low-level vision tasks\cite{wei2021drn}\cite{wei2021tsn}\cite{wei2021sarn}. Deep learning also has many applications\cite{song2020heart}\cite{hammad2020multitier} in the medical field. Many computer-aided diagnosis systems\cite{jha2022retinal}\cite{huang2021dense} based on deep learning play an important role in doctors' clinical diagnosis. 
 
At present, some algorithms and systems\cite{rajpurkar2017chexnet}\cite{ayan2019diagnosis}\cite{stephen2019efficient}\cite{jaiswal2019identifying}\cite{minaee2020deep} based on deep learning can achieve good recognition and diagnosis effect of pneumonia, but there are still some challenges and problems. The main problems are as follows: (1) The number of chest X-ray radiograph data of pneumonia is small, because it is time-consuming and laborious for doctors to label them, and the acquisition of images is time-consuming and costly, too. (2) The small amount of the data results in poor generalization and robustness of the trained model, and it is easy to overfit. (3) When training the deep learning model, the data are not fully mined and utilized, and the features extracted by the network are not quite discriminative. (4) The location information of lesions painstakingly marked by doctors is not utilized. (5) Due that the incidence of pneumonia is different, the class is extremely imbalanced, which makes the model overfitted or underfitted in certain classes, and there is a severe information overlapping between different pneumonia data samples. All of these factors have negative affect on the recognition accuracy. In this paper, aimed at solving these problems, we propose a deep learning framework named Attention-Based Contrastive Learning for Class-Imbalanced X-Ray Pneumonia Lesions Recognition (denoted as Deep Pneumonia). 


\begin{figure*}
	
	
	\subfigure[Normal]{
		\begin{minipage}[b]{0.185\textwidth}
			\includegraphics[width=3.4cm]{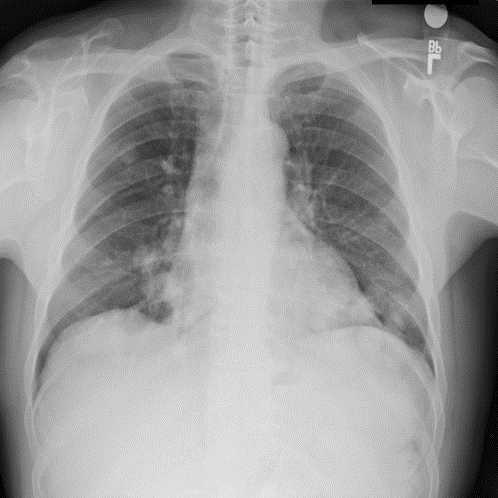}\vspace{-10pt} \\
		\end{minipage}
	}\hspace{-5pt}
	\subfigure[Pneumonia 1]{
		\begin{minipage}[b]{0.185\textwidth}
			\includegraphics[width=3.4cm]{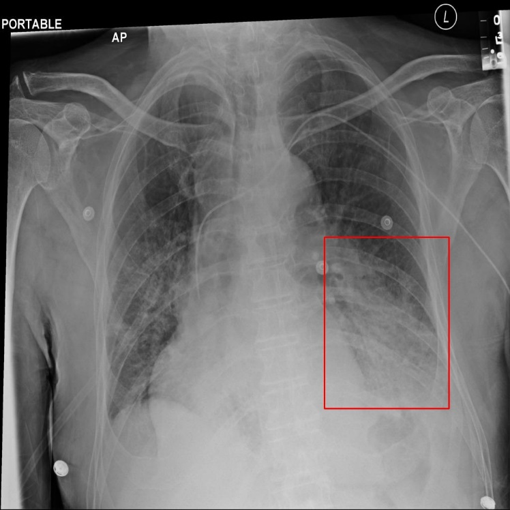}\vspace{-10pt} \\
		\end{minipage}
	}\hspace{-5pt}
	\subfigure[Pneumonia 2]{
		\begin{minipage}[b]{0.185\textwidth}
			\includegraphics[width=3.4cm]{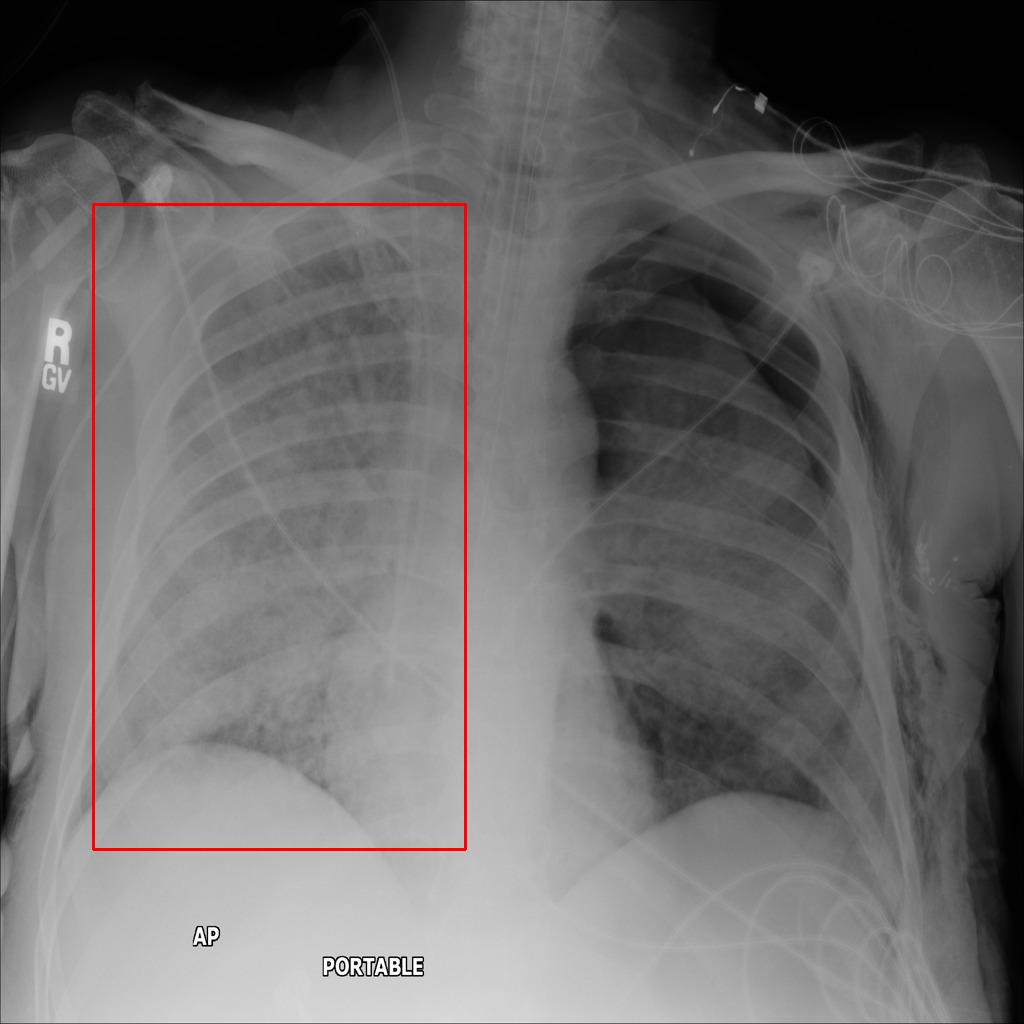}\vspace{-10pt} \\
		\end{minipage}
	}\hspace{-5pt}
	\subfigure[Pneumonia 3]{
		\begin{minipage}[b]{0.185\textwidth}
			\includegraphics[width=3.4cm]{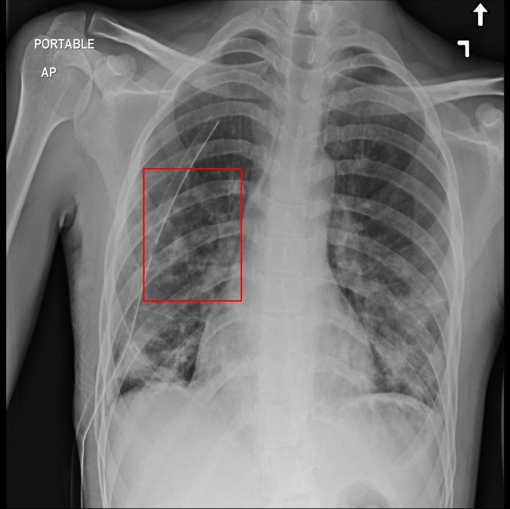}\vspace{-10pt} \\
		\end{minipage}
	}\hspace{-5pt}
	\subfigure[Pneumonia 4]{
		\begin{minipage}[b]{0.185\textwidth}
			\includegraphics[width=3.4cm]{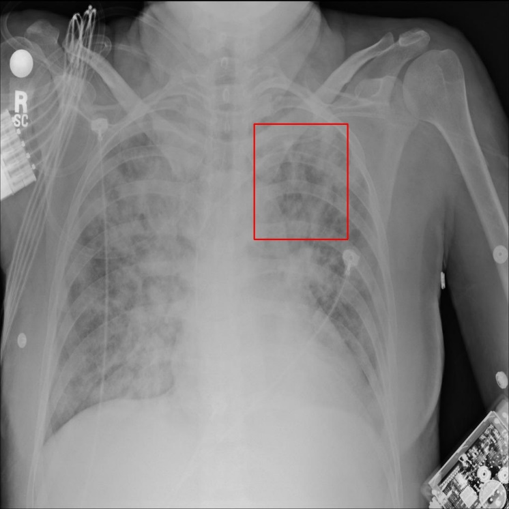}\vspace{-10pt} \\
		\end{minipage}
	}
	\caption{Examples of the five categories of X-ray images in the dataset, the first is a normal chest X-ray image without pneumonia, and the other four are images of four different types of pneumonia. The bounding boxes in the X-ray images are the area of the pneumonia lesions identified and marked by the doctor.}
	\label{dataset}
\end{figure*}

In order to solve the above-mentioned problems (1)(2)(3), we pre-train the feature extractor by adopting self-supervised contrastive learning, which can fully mine useful information from existing datasets by leveraging data augmentation and contrastive loss without using extra X-ray data. For tackling the problem (4), we introduce a mask-guided hard attention strategy and feature learning with contrastive regulation which are applied on the attention map and the extracted features respectively to guide the model to focus more attention on the lesion area identified and marked by experiened doctors. To overcome the problem (5), we adopt Class-Balanced Loss instead of traditional Cross Entropy as the loss function of classification, which can measure data overlapping and estimate the effective number of samples, then integrate a re-weighting scheme into the Focal Loss\cite{lin2017focal} for better classification. 
re-weighting scheme into the Focal Loss\cite{lin2017focal} for better classification. 
re-weighting scheme into the Focal Loss\cite{lin2017focal} for better classification. 
re-weighting scheme into the Focal Loss\cite{lin2017focal} for better classification. 
re-weighting scheme into the Focal Loss\cite{lin2017focal} for better classification. 
re-weighting scheme into the Focal Loss\cite{lin2017focal} for better classification. 

The main contributions of this paper are threefold:
\begin{itemize}
	\item In order to fully mine the existing X-ray dataset, we adopt self-supervised contrastive learning to pre-train the feature extractor without extra data;\\
	
	\item In order to leverage the location information of the lesions, we propose mask-guided hard attention with contrastive regulation module, which can guide and drive the network to focus more attention on the lesion regions;\\  
	
	\item In order to tackle the challenge of class-imbalanced dataset, we introduce Class-Balanced Focal Loss instead of Cross Entropy to re-weighting the loss weights during training.
\end{itemize}

Experiments show that these new strategies can solve those aforementioned common problems and challenges to certain extent in the field of medical image recognition.

\section{Related Works}

\subsection{Pneumonia Lesion Recognition}
There are many methods for pneumonia identification using deep learning, which have achieved good results. 
For example, CheXNet \cite{rajpurkar2017chexnet} develops an algorithm that can detect
pneumonia from chest X-rays at a level exceeding practicing radiologists.
Stephen et al. \cite{stephen2019efficient} constructs a convolutional neural network model, which is trained from scratch to extract features from given chest X-ray radiographs and classify them to determine whether a person is infected with pneumonia.
Jaiswal et al. \cite{jaiswal2019identifying} proposes a deep learning based approach for the identification and localization of pneumonia in Chest X-rays (CXRs) images.
\cite{ayan2019diagnosis} adopts Xception\cite{chollet2017xception} and Vgg16\cite{simonyan2014very} for diagnosing pneumonia with transfer learning and fine-tuning in the training stage.
Deep-covid\cite{minaee2020deep} trains four popular convolutional neural networks, including ResNet18\cite{he2016deep}, ResNet50\cite{he2016deep}, SqueezeNet\cite{hu2018squeeze}, and DenseNet-121\cite{huang2017densely}, to identify COVID-19 disease with transfer learning. Although these methods can achieve good accuracy in pneumonia recognition, they just simply apply some basic classification deep learning models to pneumonia X-ray recognition and do not solve the common problems and challenges that we mentioned above in the field of medical image recognition.

\subsection{Contrastive Learning}
In recent years, with the emergence of contrastive learning (CL) methods\cite{caron2018deep}\cite{chen2020simple}\cite{he2020momentum}\cite{wu2021contrastive}\cite{wang2021dense}, self-supervised learning\cite{misra2020self}\cite{jaiswal2021survey}\cite{jing2020self}\cite{khosla2020supervised} has attracted a lot of attention. Self-supervised learning can generate positive and negative samples through data augmentation of the anchor images, and label information can be constructed according to the pair of positive and negative samples during training, which can avoid the use of large amounts of labeled data. Such strategy is helpful to some tasks where label information is difficult to obtain.
Contrastive learning generates positive and negative sample pairs of the anchor image through two independent data augmentation, and then applies a shared-weight encoder (usually a feature extraction network, such as VGG, ResNet, etc) to extract the features of the images (encode the images into the feature space). Then the dimension of the extracted features is reduced and represented as a low-dimensional vector through the projector (usually a Multi-Layer Perception), in a metric embedding space, these vectors are constrained with contrastive loss in the embedding space. The central idea of contrastive learning is to bring similar instances closer and push away dissimilar instances far from each other by measuring the closeness between the embeddings of two samples\cite{jaiswal2021survey}.

\subsection{Attention Mechanism}
Deep learning allows networks to allocate more attention to the regions of interest (ROIs) by mimicking human attention mechanisms. Visual attention mechanisms have achieved great success in computer vision, such as image recognition\cite{woo2018cbam}\cite{park2018bam}\cite{hu2018squeeze}, pedestrian re-identification\cite{song2018mask} and image enhancement\cite{wei2021tsn}\cite{lv2021attention}. The main challenge of introducing attention mechanisms is what kind of top-down cues can be used and how to guide the bottom-up processing using these top-down cues\cite{niu2021review}\cite{zhao2019object}.

\begin{figure}[htbp]
	\flushleft
	\includegraphics[width=8.5cm]{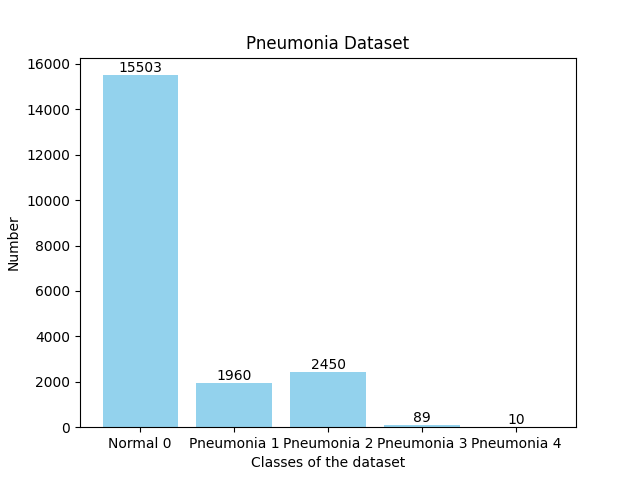}
	\caption{The distribution of the entire pneumonia dataset and the number of each category. We can see that the distribution of the dataset has a shape of long tail, which does harm to the performance of classification.}
	\label{data}
\end{figure}

\begin{figure*}[htbp]
	\centering
	\includegraphics[width=18cm]{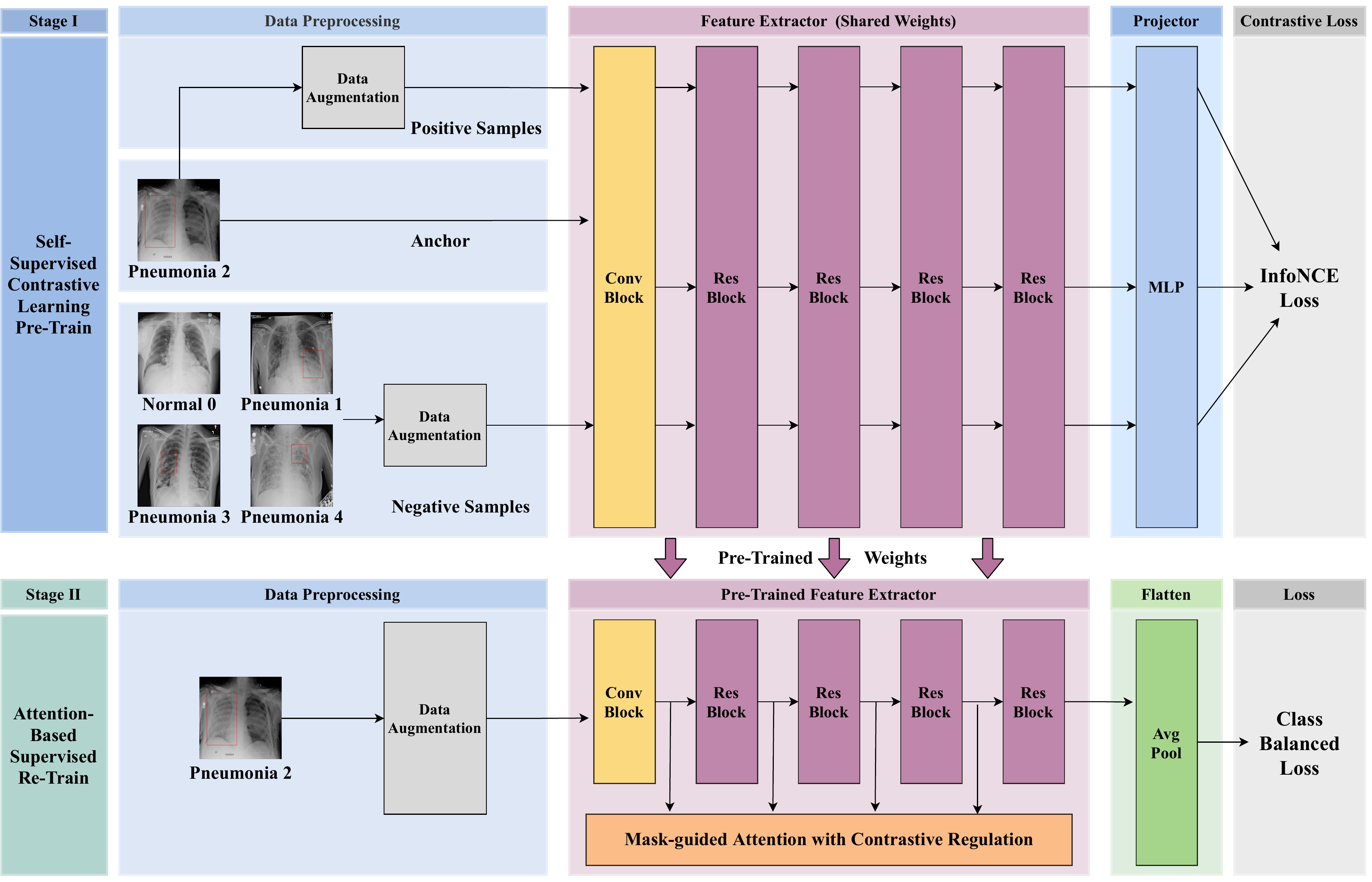}
	\caption{The framework of the proposed Deep Pneumonia. The whole pipeline includes two stages, namely Self-supervised contrastive learning pre-train stage and Attention-based supervised re-train stage. The detailed schematic of the proposed module Mask-guided Hard Attention and Feature Learning with Contrastive Regulation (MGACR) can be seen in Fig.\ref{mga}.}
	\label{model}
\end{figure*}

\section{Datasets}
The dataset we employed consists of 20,012 X-ray images of pneumonia. As shown in Fig.\ref{dataset}, there are five categories in the dataset, one category is normal X-ray images without pneumonia lesion, and the other four categories are X-ray images with different kinds of pneumonia. The pneumonia lesion areas of these four types of images are marked by highly-trained radiologists through bounding boxes, whose coordinates are available. We divided the pneumonia dataset into the training set and test set, and the rule of division is to randomly select 10\% of images of each category to form the test set, and the remaining 90\% is used as the training set. Each image is a single-channel chest X-ray with the resolution of 1024x1024. The dataset is available at https://god.yanxishe.com/23?from=god\_home\_list.

As shown in Fig.\ref{data}, the distribution of this dataset is skewed, with a long-tail. In this kind of dataset, a few dominant classes have most of the samples (for example, the samples of normal X-ray radiography without pneumonia lesions accounts for the majority of the total X-ray images), while most other classes contain relatively few examples (the incidence of some pneumonia is low, so these kinds of pneumonia are less than any other type of pneumonia in term of number. Compared with normal images without pneumonia lesions, they are even negligible). In general, models trained on such data perform poorly for weakly represented classes (a few types of pneumonia) \cite{cui2019class}.

\section{Methodology}

\subsection{Framework and Network}
As shown in Fig.\ref{model}, we propose a framework named Attention-Based Contrastive Learning for Class-Imbalanced X-Ray Pneumonia Lesion Recognition (denoted as Deep Pneumonia). There are two stages in the framework: Self-supervised contrastive learning based pre-training stage and Attention-based supervised re-training stage. 

In stage I, we adopt self-supervised contrastive learning strategy to pre-train the ResNet18\cite{he2016deep} as the feature extractor backbone. In this stage, following \cite{chen2020simple}, firstly we apply two individual data augmentation operations to the anchor image to generate the corresponding positive and negative samples. Then we feed the three kinds of samples (anchor, positive and negative samples) into the weight-shared feature extractor backbone (ResNet18\cite{he2016deep}) to extract the features of these three samples. The feature extractor can also be viewed as an encoder. Then we use Multi-Layer Perception (MLP) as the projector to reduce the dimension of encoder's output features and project these features to the low-dimensional space for representing them as metric embedding vectors. During model training in stage I, the contrastive loss\cite{chen2020simple} is used to constrain the training process of self-supervision, and the label information of samples can be constructed by the generated positive and negative sample pairs themselves without the need of given label information, so the whole process is self-supervised. After the training of stage I, we save the model and weights of the ResNet feature extractor (encoder) as the pre-trained model, which is used for feature extraction in the next stage. 

In stage II, we use the pre-trained feature extractor (ResNet18) to extract the features of the X-ray image. In this stage, we discard the MLP and adopt Adaptive Average Pooling to flatten the features into vectors. And then a fully-connected layer is used for classification.
ResNet18\cite{he2016deep} is adopted as our backbone, extracting features with a convolutional layer in the first layer, and then feeding them into the network which is composed of several residual blocks to extract deeper semantic features. There are four groups of residual layers in the network, and each residual layer is composed of one or several residual blocks\cite{he2016deep}. The number of residual blocks in each residual layer can be adjusted. After each residual layer, the scale of the image is reduced. In order to leverage the location information of the lesion area marked by doctors, so that the network can focus more attention on the lesion area and improve the classification performance, we feed the features of different scales outputted from each residual layer into the proposed module named Mask-guided Attention with Contrastive Regulation (MGACR), which can guide and drive the network to learn the expected attention map through the guidance of the given masks. In addition, the strategy of contrastive regulation is adopted in both the attention map and the features after attention weight is applied, so that the network can learn in a better way.
Previous classification methods\cite{chollet2017xception}\cite{simonyan2014very}\cite{he2016deep}\cite{huang2017densely} based on deep learning generally use Cross-Entropy (CE) as the loss function of classification, but the classes of pneumonia X-ray dataset are extremely imbalanced, therefore, we adopt Class-Balanced Focal Loss\cite{cui2019class} to replace the traditional Cross-Entropy as the loss function of classification task.

\subsection{Pre-training with Self-supervised Contrastive Learning}
There is small amount of X-ray pneumonia data, and the economic cost and time cost of obtaining the dataset are high, therefore, it is necessary to make full use of the existing dataset. In order to fully mine the existing X-ray pneumonia dataset without using external datasets, in satge I, we adopt Self-supervised Contrastive Learning\cite{chen2020simple} strategy to pre-train the feature extractor backbone. 

Before the training, positive and negative samples are generated through two individual data augmentation operations. Based on contrastive learning\cite{chen2020simple}\cite{misra2020self}\cite{he2020momentum}, we apply data augmentation to each image in the dataset. Assuming that we select an image as anchor, samples obtained by this anchor image through data augmentation are regarded as positive samples. Samples obtained through data augmentation of other images are negative samples. Then, we feed these three kinds of samples (anchor, positive and negative samples) to a weight-shared feature extractor (encoder) to extract features. We adopt ResNet18\cite{he2016deep} as the backbone of our encoder, then a Multi-Layer Perception (MLP) is used as the projector to reduce the dimension of the extracted high-dimensional features, and turn the encoder's output features into low-dimensional vectors. The features are projected to a metric embedding space. Then InfoNCE Loss\cite{chen2020simple} (a kind of contrastive loss) is used to constrain the entire training process of contrastive learning. We build the contrastive loss following the rule of maximizing the disagreement among feature vectors from different categories, minimizing the divergence among those belonging to the same category\cite{chen2020simple}\cite{xie2021detco}.
The InfoNCE Loss is defined as follow:

\begin{align}
	\mathcal L_{i,j} = -log\frac{exp(sim(z_{i},z_{j})/\tau )}{\sum_{k=1}^{2N}\left [ k\neq j \right ]exp(sim(z_{i},z_{k})/\tau)}
\end{align}
where $\tau$ = 0.2 is a temperature hyper-parameter. Sim(,) measures the similarity of two normalized vectors by dot production. $z_{i}$ is the vector of anchor sample $i$, $z_{j}$ is the vector of positive sample $j$, and $z_{k}$ is the vector of negative sample $k$.
By optimizing this contrastive loss, the model can minimize the distance between views from the same images and maximize the distance between views from different images in the latent feature space.

After the pre-training based on self-supervised contrastive learning, we only need to save the model and weights of the feature extractor backbone, and the MLP projector is discarded. The saved encoder backbone will be directly used as the feature extractor to extract image features in re-training stage.

\subsection{Mask-guided Attention with Contrastive Regulation}
Through the pre-training stage, we can obtain a powerful feature extractor that can learn and extract image features well. 
The pneumonia X-ray dataset also provides the valuable location information of the lesion areas marked by the doctor. In previous X-ray image recognition methods\cite{rajpurkar2017chexnet}\cite{stephen2019efficient}\cite{jaiswal2019identifying}\cite{minaee2020deep} of pneumonia, the location information of the lesion was not used. Inspired by the idea adopted for the task of pedestrian re-identification\cite{song2018mask}, we propose a module, namely Mask-guided Attention with Contrastive Regulation (MGACR), which consists of two strategies, i.e., Mask-guided Hard Attention and Features Learning with Contrastive Regulation, to leverage the location information of the lesion areas marked by the doctor as the priori attention information, and we add contrastive regulation on both the learning of the attention map and the features weighted by the corresponding attention map into the training process to guide and drive the network to focus more attention on the lesion regions.

\begin{figure}[htbp]
	\includegraphics[width=8.5cm]{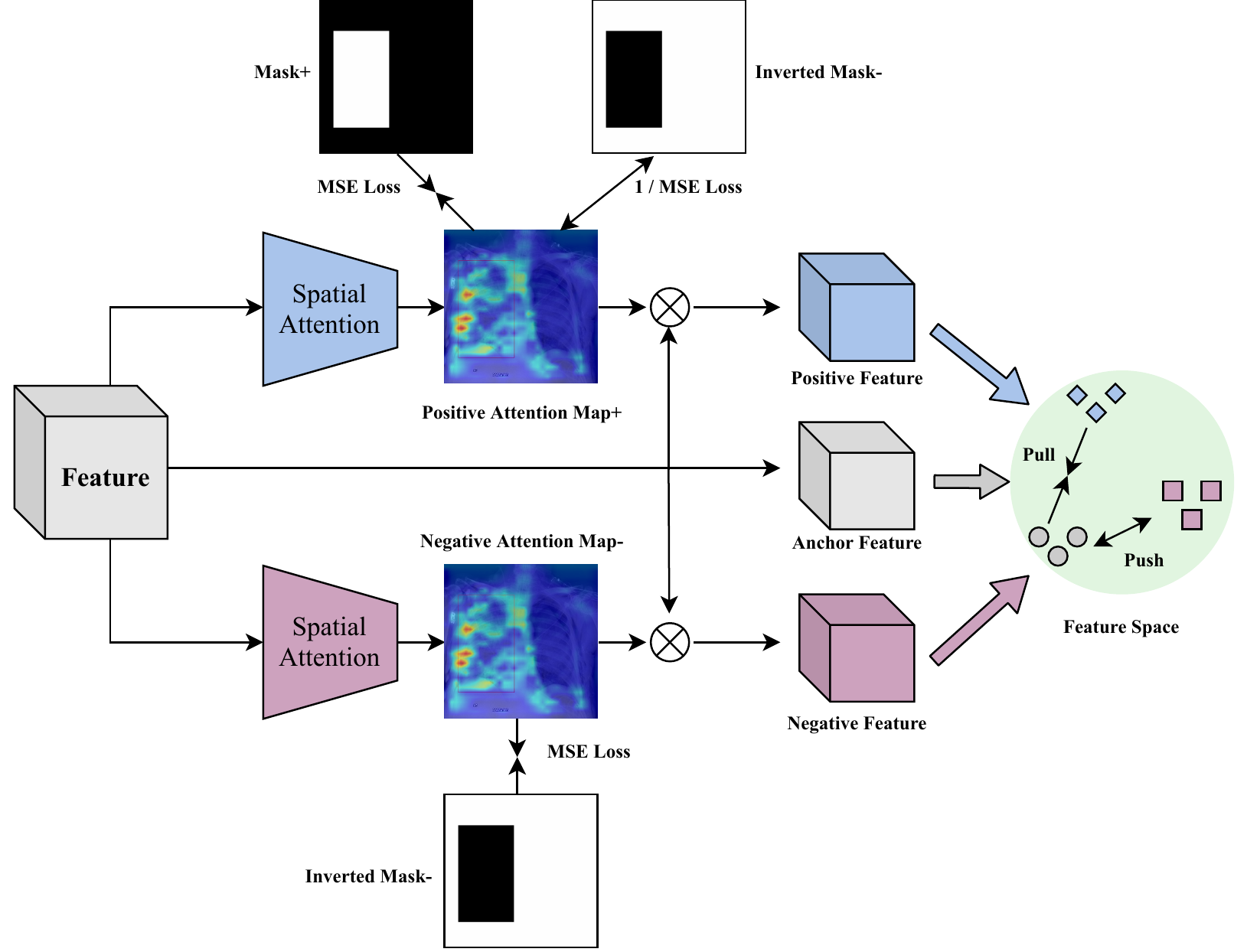}
	\caption{The schematic of the proposed Mask-guided Attention with Contrastive Regulation (MGACR) module. This module consists of two parts, one is mask-guided hard attention with contrastive regulation, and the other is feature learning with contrastive regulation. For the detailed explanation of the Spatial Attention Module, please refer to \cite{woo2018cbam}. The extracted features are constrained in the feature space by the rule of contrastive regulation.}
	\label{mga}
\end{figure}

\begin{figure}
	\subfigure[Pneumonia Image]{
		\begin{minipage}[b]{0.3\linewidth}
			\includegraphics[width=2.7cm]{pic/dataset/2.jpg}\vspace{-3pt}
	\end{minipage}}
	\vspace{-4pt}
	\subfigure[Mask+]{
		\begin{minipage}[b]{0.3\linewidth}
			\includegraphics[width=2.7cm]{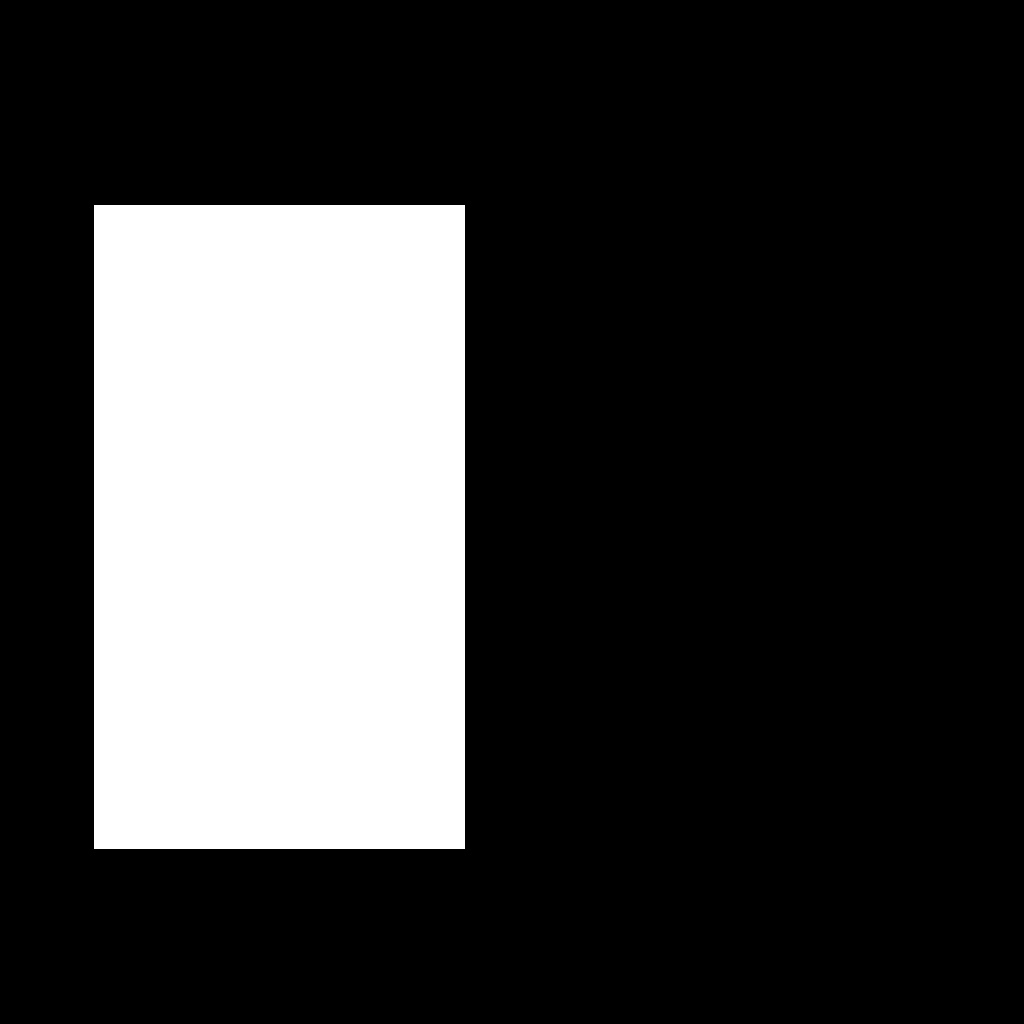}\vspace{-3pt}
	\end{minipage}}
	\vspace{-4pt}
	\subfigure[Mask-]{
		\begin{minipage}[b]{0.3\linewidth}
			\includegraphics[width=2.7cm]{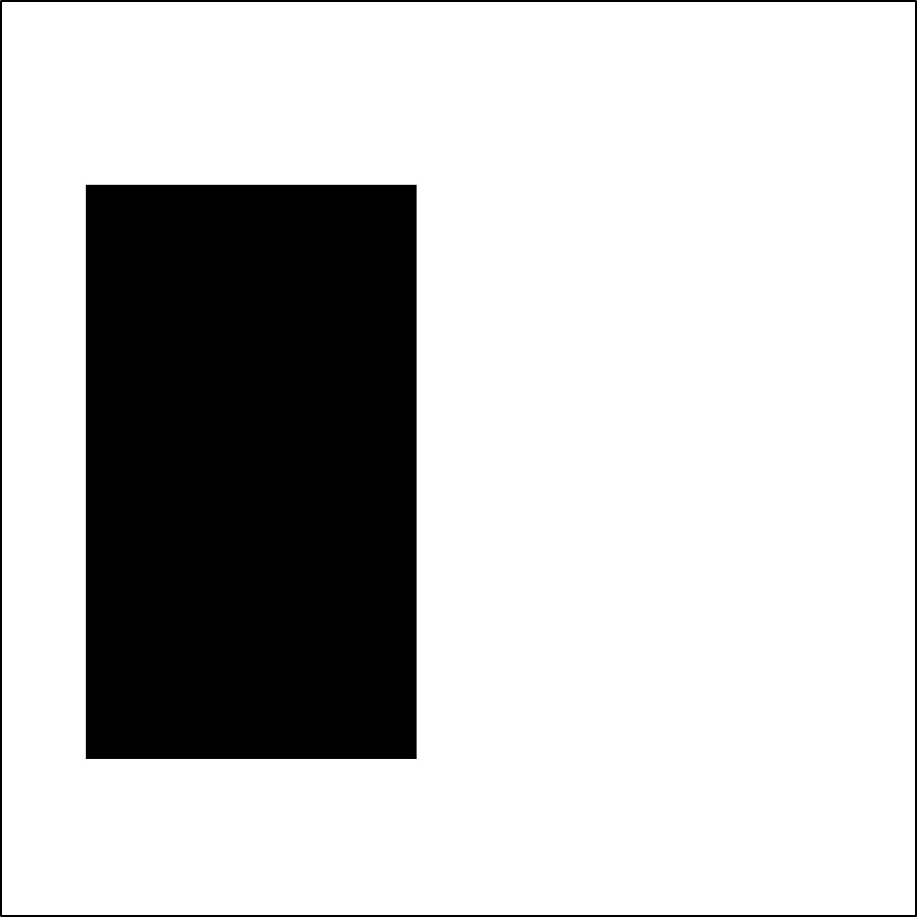}\vspace{-3pt}
	\end{minipage}}
	\vspace{3pt}
	\caption{The example of the pneumonia X-ray images with the marked lesion region via bounding box, and their corresponding positive mask (mask+) and inverted mask (mask-).}
	\label{mask}
\end{figure}

As shown in Fig.\ref{model}, in order to drive the network to focus on the lesion area, we feed the features of different scales outputted from each residual layer into the MGACR module in the re-training stage to constrain these features under the rule of contrastive regulation. We adopt the hard attention approach, that is, using the masks as the prior information to guide the learning of spatial attention and force the network to focus more attention on the discriminative lesion regions.
As shown in Fig.\ref{mask}, (a) is an X-ray image with pneumonia lesion, and the pneumonia lesion is marked by doctors through bounding boxes. According to these bounding boxes, we can generate corresponding masks and inverted masks, for example, (b) and (c) in Fig.\ref{mask}. We can guide and drive the network to pay more attention to these regions by constraining the network's attention map and the weighted features with these masks using contrastive regulation.

Next, we demonstrate in detail the working mechanism of MGACR module and how to drive the network to focus more attention on the lesion region marked with the bounding boxes by doctors. As shown in Fig.\ref{mga} and Eq.\ref{att+}, we calculate two independent spatial attention maps for the features of the input X-ray images. Then we regard one of these two attention maps as the positive attention map (map+) and use MSE Loss to constrain it with its corresponding mask (mask+). After normalization, the values of the area containing lesions in the mask are 1 and the values of other areas are 0. The purpose of such processing is to drive the distribution of weights in the positive attention map as similar as possible to the value distribution in the mask.
At the same time, we drive the positive attention map to stay away from the inverted mask (mask-) as far as possible. So, as indicated by Eq.\ref{att+}, we also calculate MSE Loss between the the positive attention map and its inverted mask, and then put the loss function on the denominator of the fraction. Similarly, as shown in Eq.\ref{att-}, MSE Loss is used to constrain another attention map and its inverted mask (mask-), and a negative attention map (map-) is obtained. The weights of this attention map are distributed outside the lesion area, which is opposite to map+.

\begin{align}
	\mathcal L_{att+} = \frac{MSE(map+,mask+)}{MSE(map+,mask-)}
	\label{att+}
\end{align}

\begin{align}
	\mathcal L_{att-} = MSE(map-,mask-)
	\label{att-}
\end{align}

The total loss function for mask-guided hard attention with contrastive regulation is as follow:

\begin{align}
	\mathcal L_{att\_cr} = \mathcal L_{att+} + \mathcal L_{att-}
	\label{att_cr}
\end{align}

We also use contrastive regulation to constrain the attention-weighted features. The specific method is as follows:: multiplying the input features with the positive attention map and negative attention map, wihch contain respectively the positive attention weights and negative attention weights, to obtain the features weighted by the positive attention and negative attention maps. According to the rule of contrastive learning\cite{chen2020simple}, we expect that the attention weight distribution of the original input features should be as similar as possible to the features integrated with positive attention weights, and at the same time dissimilar to the features weighted by negative attention.

In order to achieve this goal, we construct a loss function as shown in Eq.\ref{feat_cr}. For the anchor features, L1 Loss is used to constrain them and the two features weighted respectively by the positive and negative attention maps. In order to drive the anchor features as similar as possible to the positive features, we put the calculated loss between the anchor features and the positive features in the numerator of the fraction to learn the positive features, and in order to make the anchor feature as dissimilar as possible from the negative features and far from them, we put the calculated loss between anchor feature and negative feature in the denominator, so that the three kinds of features can form a contrastive constraint.

\begin{align}
	\mathcal L_{feat\_cr} = \frac{\mathcal L1(feat,feat*map+)}{\mathcal L1(feat,feat*map-)}
	\label{feat_cr}
\end{align}

Therefore, the total loss function for the proposed Mask-guided Attention with Contrastive Regulation (MGACR) module is as follow:

\begin{align}
	\mathcal L_{mgacr} = \mathcal L_{att\_cr} + \mathcal L_{feat\_cr}
\end{align}

Through this proposed attention module, the network can learn the latent pattern from the diagnostic experience of the doctors, and focus on the particular regions (the lesion regions) in the X-ray image where the clinician would normally focus. In a word, that is learning to see where the doctors want to see.

\begin{table*}[htbp] 
	\centering \caption{The experimental results of our model. We conducted the ablation study for all four techniques.}
	\begin{tabular}{  c | c | c  c  c  c | c  c }
		\hline  Backbone  & Image Size   & CL Pre-Train  & Mask-guided Attention  & Feature Learning with CR   & CB Focal Loss  & Accuracy (\%)  &Boost (\%)  \\
		\hline
		\hline  
		ResNet18 & 224x224 & -  & -  & -  & -  & 79.10  & - \\
		ResNet18 & 224x224 & \checkmark  & -  & -  & -  & 80.15  & 1.05 ↑ \\
		ResNet18 & 224x224 & \checkmark  & \checkmark  & -  & -  & 81.10  & 0.95 ↑ \\
		ResNet18 & 224x224 & \checkmark  & \checkmark  & \checkmark  & -  & 82.40  & 1.30 ↑ \\
		ResNet18 & 224x224 & \checkmark  & \checkmark  & \checkmark  & \checkmark  & 83.85  & 1.45 ↑ \\
		\hline\end{tabular}\vspace{0cm}
	\label{lol_real}
\end{table*}

\begin{table*}[htbp] 
	\centering 
	\caption{Comparison with other machine learning and deep learning methods.}
	\begin{tabular}{  c | c  c  c  c  c  c  c  c  c }
		\hline  Methods  & KNN  &Decision Tree   & AdaBoost  & Random Forest  & SVM\_poly     & VGG16  & ResNet18  &ResNet50  & \textbf{Deep Pneumonia}  \\
		\hline
		\hline  
		Accuracy (\%) & 75.85  &68.90 & 72.60  & 79.75  & 76.58   & 80.10  & 79.10  & 80.95  & \textbf{83.85} \\ 
		Training Time (min) & 25  &136 & 72  & 46  &  360   & 105  & 98  & 186  & \textbf{92} \\
		\hline\end{tabular}\vspace{0cm}
	\label{comparison}
\end{table*}

\subsection{Class-Balanced Focal Loss}
As mentioned before, in the pneumonia X-ray dataset, the images of normal cases without pneumonia account for the majority, while the images of other kinds of pneumonia only account for a small part, and the images of some rare pneumonia types are even negligible. In order to solve this problem of class imbalance among various categories of data in the field of medical image recognition, Class-Balanced Focal Loss\cite{cui2019class} has been introduced as the loss function of classification task in the field of computer vision during the re-training stage to replace the traditional Cross-Entropy Loss.

\begin{align}
	CE(p_{t}) = -log(p_{t})
	\label{ce}
\end{align}

\begin{align}
	FL(p_{t}) = -(1-p_{t})^{\gamma}log(p_{t})
	\label{fl}
\end{align}

As shown in Eq.\ref{ce} and Eq.\ref{fl}, Focal Loss (FL)\cite{lin2017focal} applies a modulating factor $(1-p_{t})^{\gamma }$ to the Cross-Entropy (CE) Loss in order to focus more learning on hard examples and down-weight the numerous easy negative samples. Focal Loss can successfully address the problem of class imbalance by reshaping the standard cross entropy loss such that it down-weights the loss assigned to well-classified examples. 

\begin{align}
	CB_{focal}(p_{t}) = -\frac{1-\beta }{1-\beta ^{n_{y}}}(1-p_{t})^{\gamma }log(p_{t})
	\label{cb}
\end{align}

As shown in Eq.\ref{cb}, we adopt Class-Balanced Focal Loss\cite{cui2019class} as our loss function of classification in the re-training stage. 
The original focal loss has an balanced variant, that is $(1-p_{t})^{\gamma }$. 
Class-Balanced Focal Loss introduces another factor $(1-\beta ) / (1-\beta ^{n_{y}})$ to Focal Loss based on the effective number of samples\cite{cui2019class}.

\section{Experiments}

\subsection{Implementation details}
We used the PyTorch to train our model on an Nvidia TITAN XP GPU.
The SGD optimizer was used for the training, the batch-size was set to 16 and the resolution of the image was rescaled from 1024x1024 to 224x224. We adopted ResNet18 as the backbone of our model. The number of the four residual blocks is set to [2, 2, 2, 2], which is a standard setting of ResNet18 according to \cite{he2016deep}.

\begin{figure}
	\subfigure[Loss in pre-training stage]{
		\begin{minipage}[b]{0.45\linewidth}
			\includegraphics[width=4.6cm]{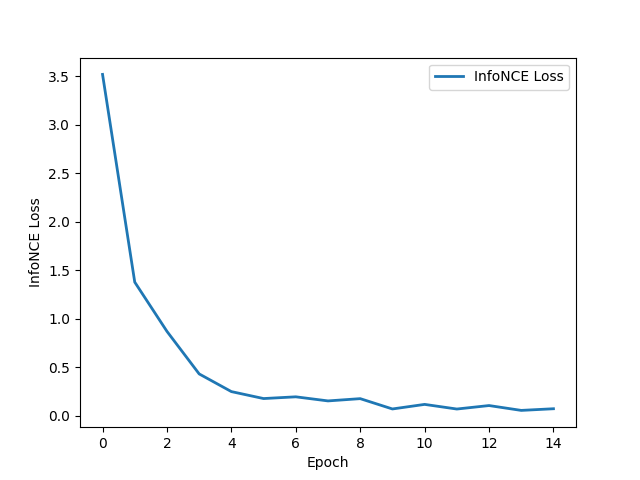}\vspace{-3pt}
	\end{minipage}}
	\vspace{-1pt}
	\subfigure[Accuracy in pre-training stage]{
		\begin{minipage}[b]{0.45\linewidth}
			\includegraphics[width=4.6cm]{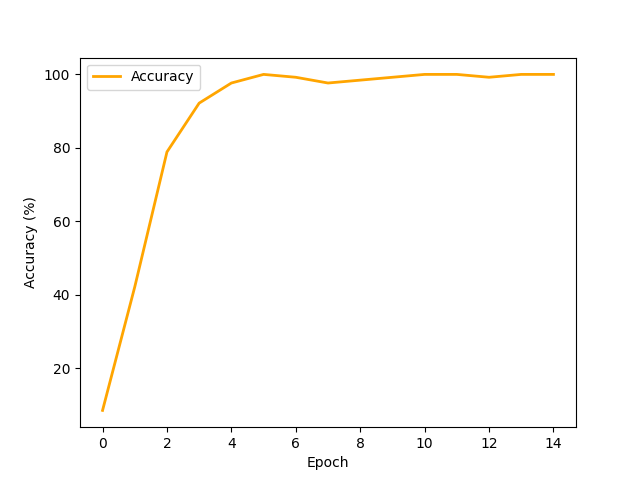}\vspace{-3pt}
	\end{minipage}}
	\vspace{-1pt}
	\caption{The change trend of loss and accuracy in self-supervised contrastive learning pre-training process.}
	\label{loss_acc}
\end{figure}

\begin{figure}[htbp]
	\includegraphics[width=9.5cm]{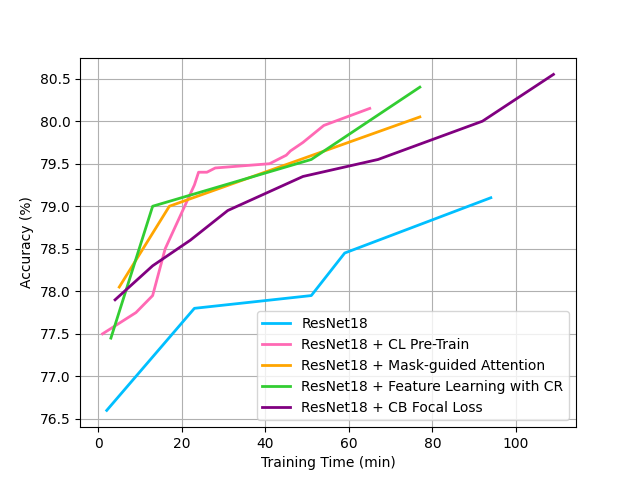}
	\vspace{-15pt}
	\caption{Ablation study of the relationship between the accuracy and training time of different techniques in the validation dataset. We tested each technique for time cost and accuracy.}
	\label{train}
\end{figure}

\begin{figure}[htbp]
	\includegraphics[width=9.3cm]{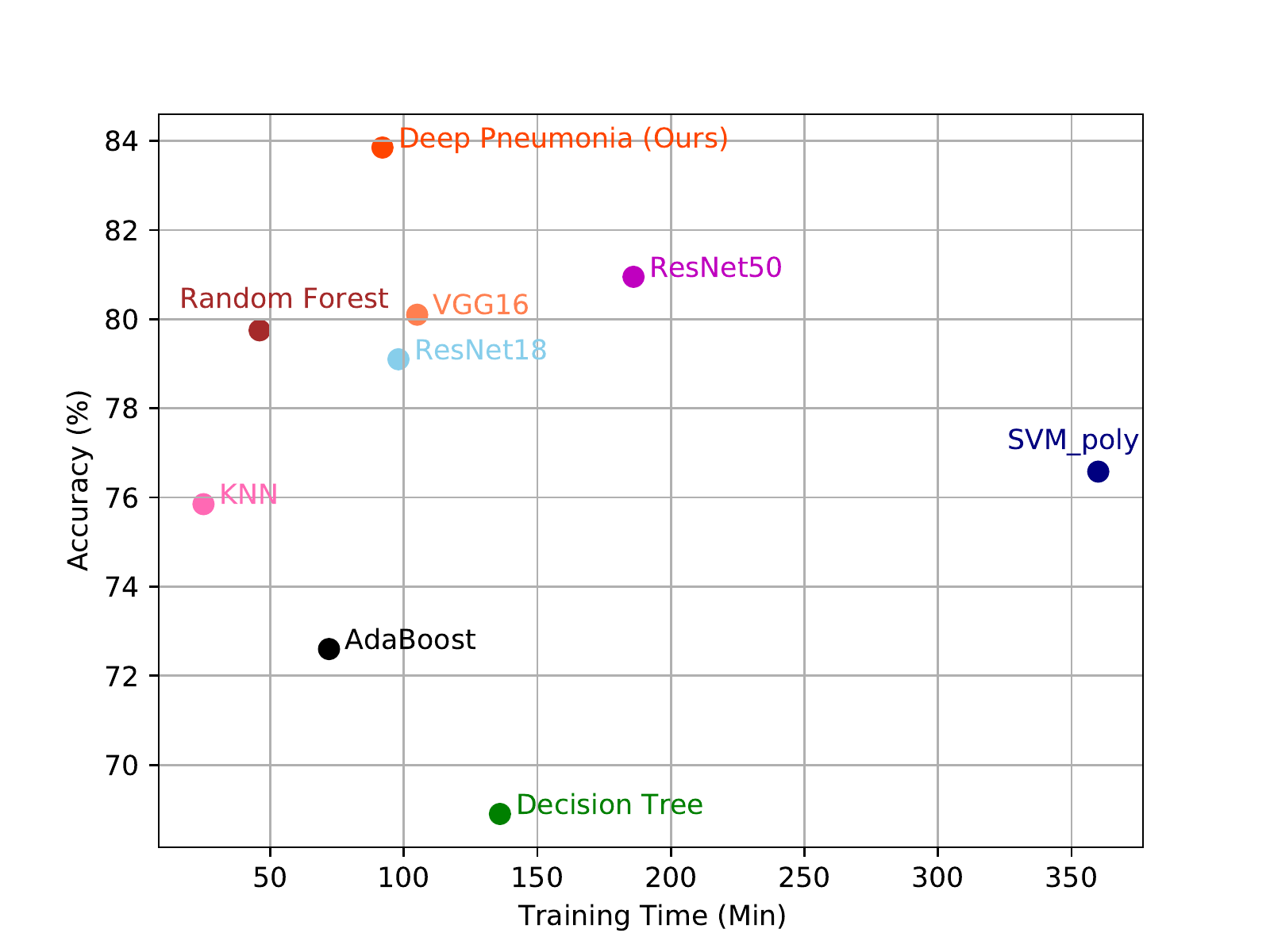}
	\vspace{-15pt}
	\caption{Comparison of training time cost and accuracy of Deep Pneumonia and other machine learning and deep learning methods.}
	\label{acc_time}
\end{figure}

\subsection{Ablation study and analysis}
As shown in Fig.\ref{loss_acc}, in the contrastive learning pre-training stage, after about 4 epochs of self-supervised training, the process of contrastive learning pre-training converged and the accuracy reached up to 100

\begin{figure}
	\subfigure[X-ray image with marked lesion]{
		\begin{minipage}[b]{0.47\linewidth}
			\includegraphics[width=4.2cm]{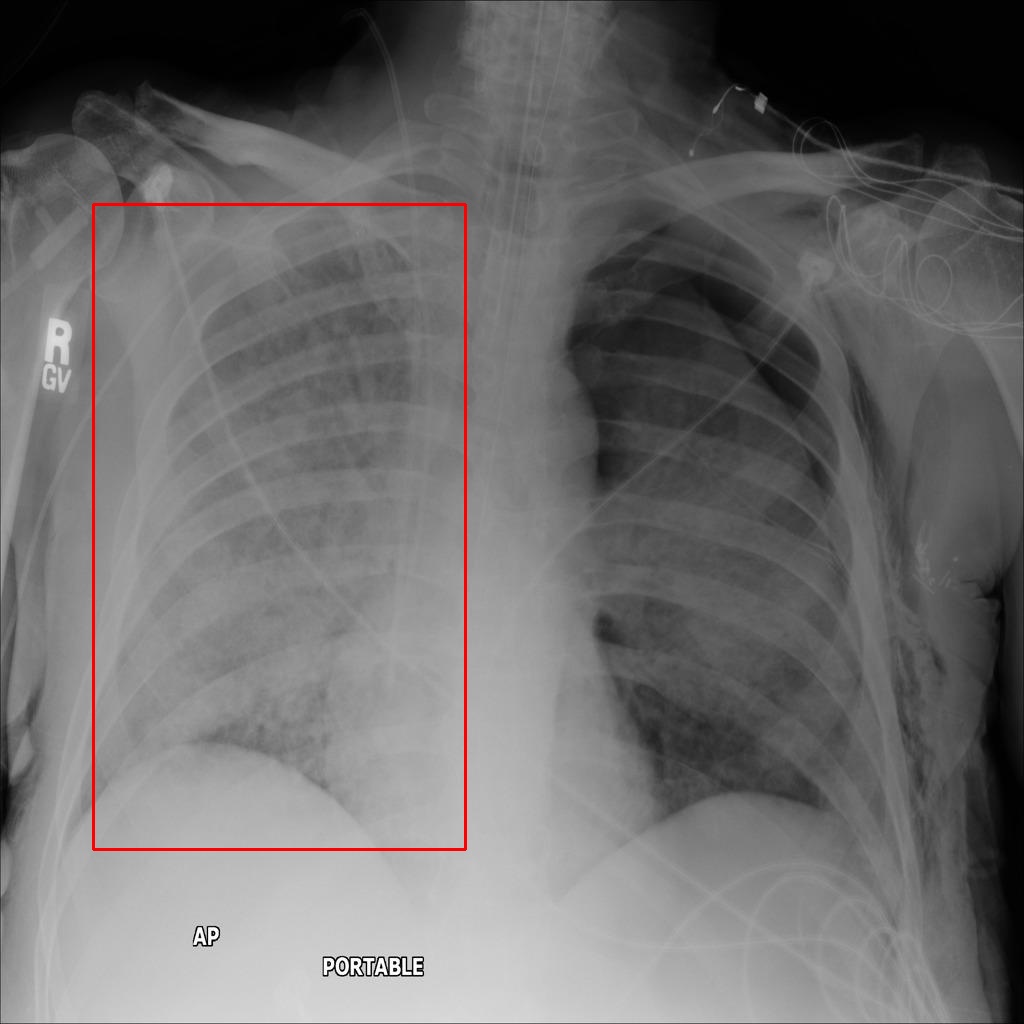}\vspace{-3pt}
	\end{minipage}}
	\vspace{-1pt}
	\subfigure[The corresponding attention map]{
		\begin{minipage}[b]{0.47\linewidth}
			\includegraphics[width=4.2cm]{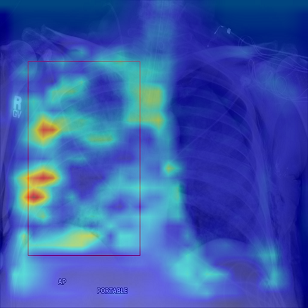}\vspace{-3pt}
	\end{minipage}}
	\vspace{-1pt}
	\caption{Visual results of attention map. We used Grad-CAM\cite{selvaraju2017grad} to visualize the activation heatmap of the model. Note that by visualizing the heat map of attention, we can detect the lesion area indirectly, and help doctors to locate the lesion area for the better diagnosis.}
	\label{abla_st}
\end{figure}

As shown in Fig.\ref{train}, we adopted ResNet18 as the backbone. We conducted ablation experiment, and added four techniques in the training process to observe the change trend of accuracy with the training time. It can be seen that the effect of training a single ResNet18 classifier without adding any techniques is worse than that of training with techniques, but the training costs less computation resources. In contrast, after adopting self-supervised contrastive learning strategy to pre-train the backbone, we found that not only the accuracy was improved, but also the convergence was much faster than that without contrastive learning pre-training. This is because after the self-supervised pre-training in the first stage, the feature extraction capability of backbone in the second stage is better than that of the backbone directly trained from scratch, so the accuracy is improved, and after the pre-training, the model converges faster during the re-training. The accuracy of the model is also improved after adding Mask-guided Hard Attention and Feature Learning with Contrastive Regulation (CR), because as shown in Fig.\ref{abla_st}, these two techniques guide and drive the network to pay more attention to the lesion regions, which contains more discriminative features. In terms of convergence speed, compared with that without Mask-guided Hard Attention and Feature Learning with CR, the network with these two techniques converges faster in the training process, because during the training, masks are used as the guidance to guide the correct training direction, and contrastive regulation is used as the constraint, which is equivalent to providing a positive pull force and a driving force for the training. All these factors accelerate the convergence speed of network training. After Class-Balanced (CB) Focal Loss was added, although the accuracy was greatly improved, the convergence speed became slower, because CB Focal Loss needs to estimate the effective number of samples during training, and then conduct the re-weighting scheme.
Note that Mask-guided Hard Attention and Feature Learning with CR only costs computation resources during the training phase, and the MGACR module can be discarded when the model conducts forward inference, which aims to help the model be better trained, so it costs no computation burden during the testing and inference phases.
In general, as shown in Table.\ref{lol_real}, these four techniques can improve accuracy very well, and the introduced extra computation burden is not too large. These four techniques have solved three common problems and challenges well in medical images and achieve promising performance.

\subsection{Comparison study}
As shown in Table.\ref{comparison} and Fig.\ref{acc_time}, we compared our method with several other machine learning and deep learning methods, the machine learning models were trained with a CPU on a high-performance server, and the deep learning models were trained on a GPU. It can be seen that our proposed method achieves good results in the accuracy, and the training time cost is also low, too.

\section*{Conclusion}
In this paper, we proposed a framework named Attention-Based Contrastive Learning for Class-Imbalanced X-Ray Pneumonia Lesion Recognition (denoted as Deep Pneumonia). In order to solve the problems of difficult acquisition of medical X-ray image data and small amount of this available data, we adopted self-supervised contrastive learning for pre-training, and fully learned and utilized existing datasets without using additional data to pre-train a powerful feature extractor. We proposed mask-guided hard attention mechanism and feature learning with contrastive regulation strategy to guide and drive the network to focus more attention on the lesion area where contains more discriminative features for solving the problem that the location information of lesions marked by doctors is not effectively utilized in pneumonia recognition. In order to solve the challenge of class-imbalanced dataset in medical image recognition, we adopted Class-Balanced Focal Loss to re-weight the loss weights during training. All in all, we hope this work can provide some heuristic ideas and inspiration to solve the common challenges in medical image recognition and help doctors better diagnose pneumonia in practice.

\bibliographystyle{unsrt}
\bibliography{ref}

\end{document}